\documentclass[preprint,5p,times,twocolumn]{elsarticle}

\usepackage{amsmath}
\usepackage{amssymb}
\usepackage{multirow}
\usepackage{graphicx}
\usepackage{color}
\usepackage{subfigure}
\usepackage{bbding} 
\newtheorem{theorem}{Theorem}
\newcommand{\ie}{\textit{i}.\textit{e}.}

\journal{Information Sciences}

\begin{document}

\begin{frontmatter}



\title{Dynamic Shuffle: An Efficient Channel Mixture Method}


\author[1]{Kaijun Gong}
\ead{18718412613@163.com}

\author[1]{Zhuowen Yin}
\ead{yinzwjohn@gmail.com}

\author[1]{Yushu Li}
\ead{eeyushuli@mail.scut.edu.cn}

\author[1,2,3]{Kailing Guo \corref{cor1}}
\ead{guokl@scut.edu.cn}

\author[1,2,3]{Xiangmin Xu}
\ead{xmxu@scut.edu.cn}

\address[1]{South China University Of Technology, Guangzhou, 510641, China}
\address[2]{Pazhou Lab, Guangzhou,  510330, China}
\address[3]{Zhongshan Institute of Modern Industrial Technology of SCUT, Zhongshan, 528400, China}

\cortext[cor1]{Corresponding author}



\begin{abstract}
The redundancy of Convolutional neural networks not only depends on weights but also depends on inputs. Shuffling is an efficient operation for mixing channel information but the shuffle order is usually pre-defined. To reduce the data-dependent redundancy, we devise a dynamic shuffle module to generate data-dependent permutation matrices for shuffling. Since the dimension of permutation matrix is proportional to the square of the number of input channels, to make the generation process efficiently, we divide the channels into groups and generate two shared small permutation matrices for each group, and utilize Kronecker product and cross group shuffle to obtain the final permutation matrices. To make the generation process learnable, based on theoretical analysis, softmax, orthogonal regularization, and binarization are employed to asymptotically approximate the permutation matrix. Dynamic shuffle adaptively mixes channel information with negligible extra computation and memory occupancy. Experiment results on image classification benchmark datasets CIFAR-10, CIFAR-100, Tiny ImageNet and ImageNet have shown that our method significantly increases ShuffleNets' performance. Adding dynamic generated matrix with learnable static matrix, we further propose static-dynamic-shuffle and show that it can serve as a lightweight replacement of ordinary $1 \times 1$ convolution.

\end{abstract}




\end{frontmatter}


\section{Introduction}
\label{Introduction}




In the past decade, we have witnessed the remarkable progress of convolutional neural networks (CNNs). CNNs have been proven to be effective and efficient in various areas espacially in vision tasks. The pursuit of improving CNNs is a constant process of leveraging model capacity within given computational resources. Benefiting from the fast development of computation hardware and sufficient data, CNNs in the early stage of the deep learning era achieved significant improvement compared to traditional machine learning methods. Because CNN models are typically over-parameterized, a well-trained network always contain too much redundancy and utilize a large amount of memory and computational power, but in return for a trivial performance increase or even decrease. 
To reduce network redundancy, later developed networks are becoming more and more compact. Shortcut connections in ResNet \cite{he2016deep} and DenseNet \cite{DBLP:conf/cvpr/HuangLMW17} promote the information transmission and make deeper and thinner neural network training become possible; group convolution \cite{DBLP:conf/cvpr/XieGDTH17} and depthwise convolution \cite{howard2017mobilenets}  significantly reduce the computation burden of convolution;
channel shuffle \cite{zhang2018shufflenet} helps the information flowing across different groups to allow more division groups.
Besides, network compression methods like pruning \cite{li2022revisiting,ding2021resrep,he2022filter,elkerdawy2022fire,liu2021discrimination,DBLP:journals/isci/FernandesY21,DBLP:journals/isci/LiZWWC22}, quantization \cite{tan2023cross,yao2022zeroquant,frantar2022optimal,li2023hard,DBLP:journals/isci/TonellottoGNGS21,DBLP:journals/isci/TonellottoGNGS21,DBLP:conf/cvpr/LiWLLLZ0T23}, matrix/tensor decomposition \cite{DBLP:conf/cvpr/YuLWT17, DBLP:journals/isci/WangZML23,DBLP:conf/bmvc/JaderbergVZ14,DBLP:conf/ijcai/XuL0WWQCLX20,DBLP:conf/cvpr/YangTW0HLLC20}, and knowledge distillation \cite{DBLP:journals/corr/HintonVD15,DBLP:journals/isci/TanL23,DBLP:conf/cvpr/YuC0J23,DBLP:conf/mm/HaoL0A021} are also proposed to remove redundant parts of existing network architectures' parameters.

However, the redundancy of CNNs not only comes from the network parameters but also comes from the feature maps. Network trimming \cite{hu2016network} have shown that a significant portion of the activations of a large network are mostly zero. The redundant parts of a model also change with different inputs. Class Activation Mapping \cite{zhou2016learning} has shown that different feature maps have different pixel-wise important regions, while Squeeze-and-Excitation Networks (SENet) \cite{hu2018squeeze} demonstrates that different channels have variant importance factors depending on changing input features.

To better deal with model redundancy, decisions have to be made dynamically regarding the input features. Dynamic compression methods \cite{wang2018skipnet,lin2017runtime} are proposed, which optionally skips model layers, channels, or adjust model width for different inputs. Specifically, dynamic channel pruning methods \cite{bejnordi2019batch,gao2018dynamic} select and skip input-dependent redundant channels of every layer according to specific criteria. However, due to their dynamic characteristics, those methods usually require dynamic memory cost to retain model performance and thus have high peak memory, which is not capable with memory-restricted hardware. 

\begin{figure*}[!t]
	\centering
	\includegraphics[width=0.98\textwidth]{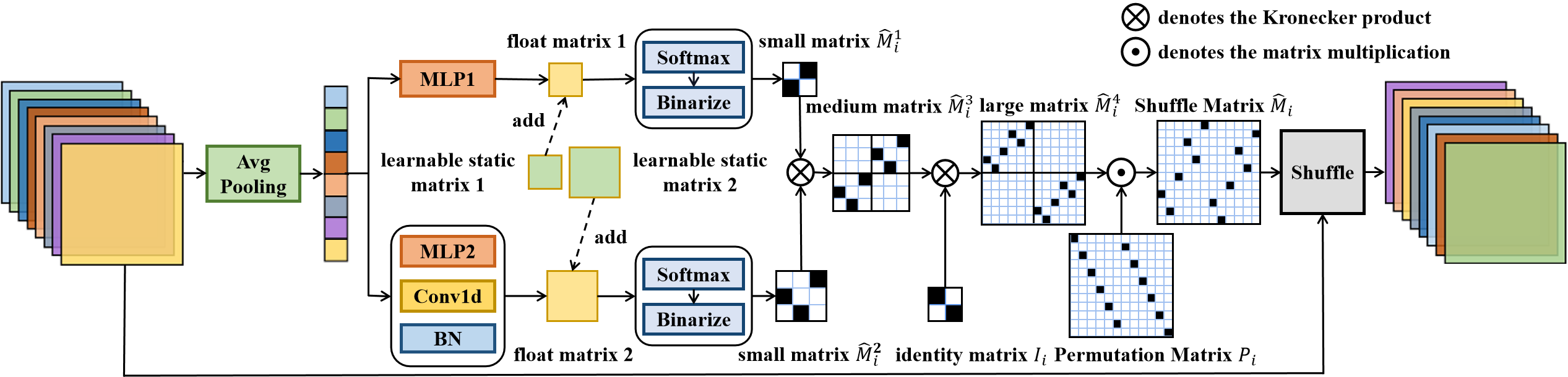}
	\caption{The structure of the Dynamic Shuffle module. If the learnable static matrices are added, it becomes the static-dynamic-mix variant, whose details can be found in Sub-Section \ref{extension} }
	\label{Fig.network_detail}
\end{figure*}

This paper tries to introduce dynamicity without drastic change of memory cost.
We design an operation called dynamic shuffle  (illustrated in Figure \ref{Fig.network_detail}), which is derived from ShuffleNet \cite{zhang2018shufflenet} but goes far from a variant.
ShuffleNet \cite{zhang2018shufflenet} uses group convolution in the first place to do intense structured pruning of the network, and then applies shuffle operation to compensate for the accuracy. The group convolution and shuffle layer is predefined in ShuffleNet, which can be viewed as a static channel mixture structure, essentially designed to reduce network redundancy by removing part of the channels. However, a static shuffle strategy can hardly remain optimal, since channel importance and redundancy shift with different input features.
In dynamic shuffle, the channel permutations can be learned like network weights and optimized based on training data. Dynamic shuffle avoids redundancy by allocating input channels in a dynamic way, which finds the suitable match of output channel for each input channel. We use a two-branch auxiliary network to generate a permutation matrix, which serve as shuffle strategies for every feature map specifically. Since the dimension of permutation matrix is proportional to the square of the number of input channels, to make the generation process efficiently, we divide the channels into groups and generate two shared small permutation matrices for each group, and utilize Kronecker product and cross group shuffle to obtain the final permutation matrices.
The auxiliary network takes globally pooled features as input, constituting only less than 1\% of the total FLOPs. 
To make the generation of permutation matrix learnable, based on our theoretical analysis, softmax, orthogonal regularization, and binarization are employed to asymptotically approximate permutation matrix. 
In general, we introduce a dynamic strategy to group convolution, thereby helping the directly and intensively pruned output channels to find their matching groups and next-layer filters with negligible computation and memory cost.

We change ShuffuleNet v1 and ShuffleNet v2 to dynamic shuffle versions and achieve considerable accuracy increase with nearly no extra computation. To show the potential application of dynamic shuffle, we use it as a lightweight replacement of $1\times 1$ convolution. Specifically, we  propose a static-dynamic-shuffle module which adds dynamic generated matrix with learnable static matrix, and directly substitute the $1\times1$ convolution layers for dimension expansion in ResNet bottleneck, sparing huge memory and computation costs and even boosting the performance.

The rest of the paper are organized as follows.
Related works of network compression methods and dynamic networks are discussed in Section 2. The dynamic shuffle mathematical formulation and the construction method of shuffle matrix are discussed in Section 3. In Section 4, we compare the performance of dynamic shuffle with baseline model and other related methods in different architectures, and ablation studies, visualization, and convergence analysis are also conducted. Section 5 concludes this paper.
\section{Related Work}\label{Related}
	\subsection{Network Compression Methods}
	To remove the redundancy of CNNs, network compression methods have been proposed, which mainly change the model statically to reduce computation and memory cost. 
	
	Quantization methods quantize the network weights and activations from float numbers into low-bit numbers. BinaryNet \cite{courbariaux2016binarized} proposed straight-through estimator (STE) to estimate gradient of quantization operation in the backward pass. STE and its variants \cite{yin2019understanding} are widely used in many network quantization methods. The accuracy of the networks reduce because of extremely low-bit. DoReFa-Net \cite{DBLP:journals/corr/ZhouNZWWZ16} implements multi-bit quantization with the round function and develops a uniform and effective quantization scheme, which becomes the basis for many subsequent quantization methods. However, such fixed mapping schemes limit the network representation ability. To enhance the performance of network quantization, learned step size quantization (LSQ) \cite{DBLP:conf/iclr/EsserMBAM20}  treats the quantizer step size as an learnable parameter, which indirectly makes the quantization range learnable. Compared with fixed mapping schemes, LSQ can better fit the changing parameter distribution by continuously changing it during the model training.
	
	Network pruning methods directly prune part of the network parameters to reduce computation and memory needs. They evaluate the importance of network parameters and remove the less important parameters depending on the compression ratio. Many pruning methods focus on measuring the importance of weights or filters. Li et al. \cite{li2016pruning} uses absolute values of weights as criteria to remove whole filters in the network with their connecting feature maps. HRank \cite{DBLP:conf/cvpr/LinJWZZ0020} applies average rank of feature map to evaluate filters' importance. Ding et al. \cite{ding2019global} select redundant parameters with Taylor series and zero the redundant parameters out during the training process gradually. Instead of measuring the importance, 
	shared channel weight convolution (SCWC) \cite{DBLP:journals/isci/TonellottoGNGS21}
	share the weights across different input channels to reduce the number of input channels for each filter through clustering and fine-tuning.
	Collaborative compression (CC) \cite{li2021towards} jointly conducts channel pruning and tensor decomposition to compress models simultaneously to have sparsity and low-rankness. 
	
	Unlike the above methods that remove relatively unimportant parts of existing networks, group convolution is originally designed as a lightweight network structure. However, within the group convolution structure, the information flow between groups is blocked, and the information representation is weakened. 
	MobileNets \cite{howard2017mobilenets} adopts $1\times1$ convolution as channel information mixture layer to compensate for the accuracy drop, while ShuffleNet \cite{zhang2018shufflenet} substitutes the $1\times1$ convolution with shuffle operation to further remove the computation and memory required by $1\times1$ convolution.	
	 AutoShuffleNet \cite{lyu2020autoshufflenet} learns the shuffle strategy of each layer based on ShuffleNet, by training to obtain a static transformation matrix for each layer. Fully Learnable Group Convolution (FLGC) \cite{wang2019fully} learns to assign channels for different groups, which results in unequal channel numbers in each group.
	Dynamic Grouping Convolution (DGConv) \cite{zhang2019differentiable} determines both group numbers and channel connections through learning. These models are trying to design an optimal static network structure for group convolution either by manual setting or learning, but without considering the relationship between network structures and current inputs.
	
    \subsection{Dynamic Networks}
	Since different network inputs have different data-dependent computational characteristics, dynamic networks are proposed to adapt the network mapping function dynamically to the referred input data \cite{han2021dynamic}. Bolukbasi et al. \cite{bolukbasi2017adaptive} set several network modules as selective components and use selection network to evaluate the appropriate modules to be used, regarding the current input data. 
	SENet \cite{hu2018squeeze} leverages feature-driven attention on the input feature map of each network layer with fully connected (FC) layers, serving as soft dynamic channel selection. Efficient channel attention (ECA) \cite{wang2020eca} introduces 1D convolution (Conv1D) to replace the FC layer in SENet for acceleration.
	Conditionally parameterized convolution (CondConv) \cite{yang2019condconv} learns specialized convolutional kernels for each sample as a linear combination of several kernels with sample dependent coefficients. Rethinking SENet as acting on the weight, WeightNet \cite{ma2020weightnet} generalizes the SENet and CondConv by simply adding one more grouped fully-connected layer to the attention activation layer.
	Dynamic Group Convolution (DGC) \cite{su2020dynamic} uses a saliency guided channel selector to adaptively select which part of input channels to be connected within each group for individual samples on the fly. These networks focus on adapting network structure/parameters with auxiliary modules dynamically according to the input features, but the outputs of their auxiliary modules need multiplication and adding operations to generate network weights.
	
	Instead of generating  weights, some methods focus on designing dynamic activation functions. Dynamic ReLU (DyReLU) \cite{chen2020dynamic} adopts piecewise linear function as activation function and generates its hyperparameters by encoding the global context similar to SENet. Unlike DyReLU, ACtivateOrNot (ACON) \cite{ma2021activate}  learns to determine the activation to be linear or non-linear with a switching factor explicitly conditioned on the input. MicroNet \cite{li2021micronet} strengthens the connection between groups by linearly combining circular shifting channels with sample dependent coefficient, and utilizes the maximum of several fusions as activation. Dynamic activation methods are orthogonal to dynamic structures. In this paper, we focus on the dynamic structure.
	
    Dynamic pruning is a dynamic counterpart of network pruning that evaluating network redundancies and prune them dynamically. For channel pruning methods, random channel pruning \cite{li2022revisiting} is a simple, general and explainable baseline, which performs well and can be used as a benchmark when evaluating more complex pruning methods. Runtime neural pruning \cite{lin2017runtime} utilized reinforcement learning to generate dynamic channel selection during inference. Bejnordi et al. \cite{bejnordi2019batch} set a residual block and used the batch shape method to gate the output channels. Selective allocation of channels \cite{jeong2019training} trains the layer to dynamically select important channels to redistribute their parameters. Manifold regularized dynamic pruning (ManiDP) \cite{tang2021manifold} focused on the inherent correlations within training datasets, embedding the manifold information of all instances into the network and conducting dynamic pruning thereupon. Despite high theoretical compression rates, these dynamic methods make task-dependent pruning decisions, thus usually requiring different memory for different inputs. Regarding this problem, feature boosting and supression (FBS) \cite{gao2018dynamic} selects and retains the first k rated filters during inference to keep a constant compression rate with a feature selection network. In our work, regarding ``shuffle" as the channel mixture method with minimal computation cost, we propose dynamic shuffle, which generates the optimal shuffle strategy to leverage the network representation power.

\section{Method}
%

\subsection{Shuffle Matrix}\label{shuffmat}

Group convolution is widely used to reduce the computation of CNNs but separates the connections of channels in different groups. ShuffleNet \cite{zhang2018shufflenet} interleaves information between groups by shuffling the channel groups manually. AutoShuffleNet \cite{lyu2020autoshufflenet} learns the shuffle strategy by training, but it is still predefined for inference.

Because the relationship of feature maps in different channels depends not only on the network weights but also on the inputs, predefining the relationship is improper. A more effective way to reorganize channels would need feature map information. Hence, we shuffle the channels dynamically depending on the input feature maps of each layer.

By reshaping each feature map into a vector, the feature maps of the $i_{th}$ layer of a CNN can be represented by $F_{i}\in\mathbb{R}^{C_{i}\times N_{i}}$, where $C_{i}$ is the channel number and $N_{i}$ is the element number of a feature map. Shuffle is an operation that rearranges the order of feature map channels in a layer, which can be implemented by memory shifting. However, it is a discrete operation that can not be directly incorporated into network training. To make the shuffle operation learnable, we reformulate it by matrix multiplication as follows:
\begin{equation}\label{shuffleeq}
F'_i=M_i F_i,
\end{equation}
where $F'_i$ are feature maps after shuffling and $M_i$ is a permutation matrix, \ie, a square binary matrix that has one entry of 1 in each row and each column and zeros elsewhere. With this notation, all shuffle operations can be represented by matrix multiplications.
%

Dynamic shuffle is to generate a permutation matrix with the current feature maps. We need to address two problems: 1) generate a matrix from feature maps, 2) make sure that the generated matrix is a permutation matrix. By using $p_{i}(\cdot)$ to denote the generation function of the $i_{th}$ layer, we have: 
\begin{equation}\label{genmateq}
M_i={p_i}(F_i).
\end{equation}
Eq. (\ref{genmateq}) can be implemented by a small auxiliary network. We leave network architecture discussion later, and show how to ensure $M_{i}$ to be a permutation matrix first.

\subsection{Loss Function}
A permutation matrix is a binary matrix with the sum of each row and each column to be one. Intuitively, we can add these constraints directly to the loss function. However, the optimization of such a zero-one programming-like problem is too complex. To use the popular stochastic gradient descent method, we try to change the constraints to optimization friendly ones. Considering that a permutation matrix is an orthogonal matrix, we utilize orthogonal constraint here as a proxy approach. With the orthogonal constraint, the simultaneous constraints of both rows and columns summing one can be replaced by either each rows summing one or each columns summing one.
It comes from the following theorem. 

\begin{theorem}
	A matrix $M\in \mathbb{R}^{C\times C}$ with its element in the $i_{th}$ row and $j_{th}$ column being $m_{ij}$ is a permutation matrix if it satisfies the following conditions: (1) $m_{ij}\geq 0 ,\forall i,j\in[1,C]$. (2) $\sum_{k=1}^{C}m_{ik}=1, \forall i\in[1,C]$. (3) $M$ is an orthogonal matrix.
    \label{theo1}
\end{theorem}

\emph{Proof.}
Since $M$ is an orthogonal matrix, we have $MM^{T}=I$. Therefore, $\sum_{k=1}^{C}m_{ik}^{2}=1$. Since $m_{ij}\geq 0$, we have
\begin{eqnarray*}
\begin{aligned}
(\sum_{k=1}^{C-1}m_{ik})^2+m_{iC}^2&=\sum_{k=1}^{C-1}m_{ik}^{2}+2\sum_{k=1}^{C-1}m_{ik}\sum_{d\neq k,C}m_{id}+m_{iC}^2\\
&\geq\sum_{k=1}^{C-1}m_{ik}^{2}+m_{iC}^2\\
&=1.
\end{aligned}
\end{eqnarray*}
We thus get $(1-m_{iC})^{2}+m_{iC}^{2}\geq 1$. On the other hand, the maximum value of $(1-m_{iC})^{2}+m_{iC}^{2}$ is 1 when $0\leq m_{iC}\leq 1$. Therefore, $(1-m_{iC})^{2}+m_{iC}^{2}=1$, whereas  solutions to the equation are 1 and 0. When $m_{iC}=1$, we will have $\sum_{k=1}^{C-1}m_{ik}=1-m_{iC}=0$ and thus $m_{ik}=0 ,\forall k \in[1,C-1]$. When $m_{iC}=0$, we will have $\sum_{k=1}^{C-1}m_{ik}=1-m_{iC}=1$. By similar deduction, we can conclude that for each row of $M$, there is only one entry of 1 and the other entries are zeros. If there is one column that has more than one entry of ones, the corresponding rows must be the same and the rank of $M$ is less than $C$. This contradicts to Condition (3). Therefore, $M$ is a permutation matrix.\\

The first two conditions of Theorem \ref{theo1} can be satisfied by applying softmax operation to each row of a matrix without adding constraints to the loss function. Suppose there is an auxiliary network denoted by $f_{i}$ that generates a matrix $f_i(F_i)$. Then, the generation function $p_i(\cdot)$ is given by
\begin{equation}
p_{i}(F_i)=\text{binarize}(\text{softmax}(f_i(F_i))).
\end{equation}
The function $\text{binarize}(\cdot)$ is to set the element of the maximum value in each row of the matrix to one and set the others to zeros. It is to guarantee that $M_{i}$ is for selection but not for weighted summing. To deal with the non-differentiability of $\text{binarize}(\cdot)$, we use STE \cite{courbariaux2016binarized} to approximate its gradient for training, that we only propagate gradients of the elements with value one, and cancel the gradients of zero-value elements. 

We resort to adding a regularization term to satisfy the third condition. The regularization term is given by
\begin{equation}
R(\hat{M_{i}})=\|\hat{M_{i}}\hat{M_{i}}^{T}-I\|_{F},
\end{equation}
where $\hat{M_{i}}$ denotes $\text{softmax}(f_i(F_i))$ and $I$ is an identity matrix. Suppose the layer number is $l$ and the network classification loss is $L(\mathcal{W})$, where $\mathcal{W}$ is the weights of the whole network. Our training process is to minimize the following objective function
\begin{equation}
\mathcal{L}(\mathcal{W})=L(\mathcal{W})+\lambda \sum_{i=1}^{l}(R(\hat{M_{i}})),
\end{equation}
where $\lambda\geq 0$ is a trade-off parameter. By minimizing this objective function, we gradually optimize $\hat{M_{i}}$ to satisfy the three conditions in Theorem \ref{theo1}, so that it approximates a permutation matrix. 
\subsection{Auxiliary Network Architecture}
The auxiliary network takes stem network feature maps as input and outputs a shuffle matrix. We aim to design an auxiliary network that brings negligible extra computation. 
Following the pioneering dynamic network SENet\cite{hu2018squeeze}, global average pooling, which significantly reduces the feature dimension, is utilized to obtain the input vector of the auxiliary network. 

After producing a feature vector for a layer with $C$ channels via pooling, the auxiliary network needs to generate a $C\times C$ matrix from this $C$-dimensional vector. In this case, the output dimension is significantly larger than the input dimension, especially for large channel numbers, which brings a large computational burden and may make the training unstable.

To tackle the dimension increase problem, we can divide the output channels into $g$ groups and permute the feature maps within a group via a sharing permutation matrix among all groups. 
Thus, the size of permutation matrix changes from $C\times C$ to $(C/g)\times (C/g)$. To further reduce the generated dimensions, we try to generate a large matrix via small matrices. Note that the Kronecker product of two permutation matrices is still a permutation matrix. We can thus use the Kronecker product of two small permutation matrices to replace the large permutation matrix. 
In practice, we first generate two small matrices $\hat{M}_{i}^{1}$,  $\hat{M}_{i}^{2}$ with a two-branch auxiliary network for a channel group. Then for the second step, we share the permutation matrix $\hat{M}_{i}^{1}\otimes\hat{M}_{i}^{2}$ for each group. The sharing operation is implemented by a Kronecker product between an identity matrix ${I}_{i}\in \mathbb{R}^{g\times g}$ and the result of $\hat{M}_{i}^{1}\otimes\hat{M}_{i}^{2}$. 
To increase the diversity of permutation patterns in different groups for stronger representation ability, in the third step, we mix the columns in different groups of the matrix ${I}_{i}\otimes(\hat{M}_{i}^{1}\otimes\hat{M}_{i}^{2})$ via the manual shuffle operation used in ShuffleNet \cite{zhang2018shufflenet}. Suppose the corresponding permutation matrix of the manual shuffle operation is $S$. The final permutation matrix $\hat{M}_{i}$ is given by
\begin{equation}\label{four_mat}
\hat{M}_{i}=S({I}_{i}\otimes(\text{binarize}(\hat{M}_{i}^{1})\otimes\text{binarize}(\hat{M}_{i}^{2}))).
\end{equation}
The Kronecker products of binary matrices can be implemented via memory duplication, and shuffling with $S$ is a memory shifting operation. Thus, the extra computation cost bring by Eq. (\ref{four_mat}) is negligible. 

To balance the computation burden of the two branches in the auxiliary network, the sizes of the $\hat{M}_{i}^{1}$ and $\hat{M}_{i}^{2}$ are set to near $\sqrt{C/g}\times \sqrt{C/g}$. Thus, the output dimension of each branch is the same as or near the input dimension. Since the Kronecker product is non-commutative, the small matrices play different roles. $\hat{M}_{i}^{1}$ is to determine which part to be selected coarsely, and $\hat{M}_{i}^{2}$ is to determine which channel to be selected finely. The architecture of the auxiliary network is shown in Figure \ref{Fig.network_detail}. An average pooling layer is used to reduce the feature maps into a vector. The reduced vector after average pooling is the shared input of the two branches. For the branch of coarse determination whose output is denoted by $\hat{M}_{i}^{1}$, we use a multi-layer perceptron (MLP) with two FC layers. For the branch of fine determination, we meticulously design the architecture of the branch of $\hat{M}_{i}^{2}$ since it is expected to obtain accurate position information. $\hat{M}_{i}^{2}$ is generated by a MLP with two FC layers followed by a Conv1D layer and a batch normalization (BN) layer. The Conv1D layer motivated by ECA \cite{wang2020eca} is used to extract features more efficiently, and BN is to calibrate features' distribution since the dynamic network introduces a more variety of features.


The loss function is changed into 
\begin{equation}
\mathcal{L}(\mathcal{W})=L(\mathcal{W})+\lambda \sum_{i=1}^{l}(R(\hat{M}_{i}^{1})+R(\hat{M}_{i}^{2})).
\end{equation}

Following the original ShuffleNets \cite{zhang2018shufflenet,ma2018shufflenet}, the proposed dynamic shuffle operations are implemented in stages 2, 3, and 4 of ShuffleNet v1 and v2. The details of the architectures for generating the small permutation matrices are shown in Table \ref{table_matrix}.
For ShuffleNet v1, we set the channel group number to the group number of the corresponding group convolution and use the two small matrices $\hat{M}_{i}^{1}$ and $\hat{M}_{i}^{2}$ directly to produce the permutation matrix. For ShuffleNet v2, since the feature maps are concatenated from two pathways, we only apply dynamic shuffle to the main pathway with the group number of one.
{However, for ShuffleNet v2, due to their incompatible stage widths, we first calculate the Kronecker product of $\hat{M}_{i}^{1}$ and $\hat{M}_{i}^{2}$, and clip the  slightly larger matrix into the exactly desired size with the stage width.} 


\begin{table*}[t!]
	\centering
    \small
	\caption{The dynamic shuffle module's architectures in each dynamic shuffle model. For Conv1D, c denotes the output channel number, s denotes the stride, and p denotes the padding.}
	\setlength{\tabcolsep}{1.1pt}
	\begin{tabular}{c|c|c|c|c|c|c|c|c|c}
	    \hline
		\multirow{2}{*}{Network}&\multicolumn{3}{c|}{Stage2} &\multicolumn{3}{c|}{Stage3} &\multicolumn{3}{c}{Stage4} \\
		 \cline{2-10}
		&{MLP1}&{MLP2}&{Conv1D}&{MLP1}&{MLP2}&{Conv1D}&{MLP1}&{MLP2}&{Conv1D}\\
	    \hline
        \multirow{3}{1.9cm}{\centering  ShuffleNet v1\\ ($1\times$, g=3)}&\multirow{3}{0.85cm}{\centering 60$\times$5\\5$\times$16}&\multirow{3}{0.85cm}{\centering 60$\times$5\\5$\times$20}&\multirow{3}{1.5cm}{\centering1$\times$6\\c=5, s=4\\p=1}&\multirow{3}{1.05cm}{\centering 120$\times$10\\10$\times$25}&\multirow{3}{1.05cm}{\centering 120$\times$10\\10$\times$40}&\multirow{3}{1.5cm}{\centering 1$\times$13\\c=8, s=5\\p=4}&\multirow{3}{1.05cm}{\centering 240$\times$20\\20$\times$16}&\multirow{3}{1.05cm}{\centering 240$\times$20\\20$\times$80}&\multirow{3}{1.5cm}{\centering 1$\times$26\\c=20, s=4\\p=11}\\
        &&&&&&&&&\\
        &&&&&&&&&\\
        \hline
        \multirow{3}{1.9cm}{\centering  ShuffleNet v1\\ ($1\times $, g=8)}&\multirow{3}{0.85cm}{\centering 96$\times$3\\3$\times$16}&\multirow{3}{0.85cm}{\centering 96$\times$3\\3$\times$12}&\multirow{3}{1.5cm}{\centering 1$\times$4\\c=3, s=4\\p=0}&\multirow{3}{1cm}{\centering 192$\times$6\\6$\times$16}&\multirow{3}{1.05cm}{\centering 192$\times$6\\6$\times$24}&\multirow{3}{1.5cm}{\centering 1$\times$8\\c=6, s=4\\p=2}&\multirow{3}{1.05cm}{\centering 384$\times$12\\12$\times$16}&\multirow{3}{1.05cm}{\centering 384$\times$12\\12$\times$48}&\multirow{3}{1.5cm}{\centering 1$\times$16\\c=12, s=4\\p=6}\\
        &&&&&&&&&\\
        &&&&&&&&&\\
        \hline
        \multirow{3}{1.9cm}{\centering  ShuffleNet v2\\ (1$\times$)}&\multirow{3}{0.85cm}{\centering 58$\times$3\\3$\times$36}&\multirow{3}{0.85cm}{\centering 58$\times$7\\7$\times$60}&\multirow{3}{1.5cm}{\centering 1$\times$20\\c=10, s=6\\p=7}&\multirow{3}{1.05cm}{\centering 116$\times$7\\7$\times$36}&\multirow{3}{1.05cm}{\centering 116$\times$15\\15$\times$120}&\multirow{3}{1.5cm}{\centering 1$\times$40\\c=20, s=6\\p=17}&\multirow{3}{1.05cm}{\centering 232$\times$14\\14$\times$36}&\multirow{3}{1.05cm}{\centering 232$\times$29\\29$\times$234}&\multirow{3}{1.5cm}{\centering 1$\times$39\\c=40, s=6\\p=36}\\     
        &&&&&&&&&\\ 
        &&&&&&&&&\\
        \hline
        \multirow{3}{1.9cm}{\centering  ShuffleNet v2 \\ (1.5$\times$)}&\multirow{3}{0.9cm}{\centering 88$\times$5\\5$\times$81}&\multirow{3}{0.85cm}{\centering 88$\times$11\\11$\times$90}&\multirow{3}{1.5cm}{\centering 1$\times$30\\c=10, s=9\\p=11}&\multirow{3}{1cm}{\centering 176$\times$11\\11$\times$81}&\multirow{3}{1.05cm}{\centering 176$\times$22\\22$\times$180}&\multirow{3}{1.5cm}{\centering 1$\times$60\\c=20, s=9\\p=26}&\multirow{3}{1.05cm}{\centering 352$\times$22\\22$\times$81}&\multirow{3}{1.05cm}{\centering 352$\times$45\\45$\times$360}&\multirow{3}{1.5cm}{\centering 1$\times$120\\c=40, s=9\\p=56}\\ 
        &&&&&&&&&\\
        &&&&&&&&&\\
		\hline
	\end{tabular}
\label{table_matrix}
\end{table*}

\subsection{Channel Information Iteration}\label{extension}
Shuffle operation can be implemented via $1\times1$ convolution. We will show that dynamic shuffle can go far from variants of ShuffleNets by serving as a lightweight replacement of $1\times1$ convolution, which is widely used for channel information iteration and occupies most computation in modern CNNs. Here we takes the popular ResNet \cite{he2016deep} as an example.
The bottleneck structure in ResNet stacks three layers of $1\times 1$, $3\times 3$, and $1\times 1$ convolutions, in which $1\times 1$ convolution is used for dimensionality reduction or expansion. Shuffle operation can be considered as selecting an optimal input channel for each output channel. If $1\times 1$ convolution reduces the dimension, replacing it with shuffling will select part of the input channels and lose some useful information. Thus, we only use dynamic shuffle to substitute $1\times1$ convolution that does not reduce channel dimension. 
Compared to $1\times1$ convolution, the dynamic shuffle module only takes a vector as input for the auxiliary network and needs much less computation. In practice, channel shuffle could be replaced by memory shifting during inference, where the computation cost is largely reduced. It is to be noted that for channel expansion, the shuffle matrix here becomes a rectangle matrix, where orthogonal regularization term is not applicable. We need to make sure that each matrix row has only one entry of 1. From the proof of Theorem \ref{theo1}, we can deduce that this can be achieved if the square sum of a row is 1, with softmax operation. Thus, we change the regularization term to
\begin{equation}
R(\hat{M_{i}})=\sqrt{\sum_{j}(\|\hat{M}_{i,j}\|_{2}-1)^2},
\end{equation}
where $\hat{M}_{i,j}$ denotes the $j_{th}$ row of $\hat{M}_{i}$.

To obtain a more stable outcome for the rectangular matrix, we substitute the dynamically generated permutation matrix  with the sum of a static learnable matrix and the dynamically generated matrix, as is shown with dotted line in Figure \ref{Fig.network_detail}. The new module enhancing the model stability under more complex circumstances is called static-dynamic-mix shuffle.

\section{Experiments}
We apply dynamic shuffle to ShuffleNet \cite{zhang2018shufflenet}, ShuffleNet v2 \cite{ma2018shufflenet} and ResNet-50 \cite{he2016deep}, and conduct experiments on CIFAR-10, CIFAR-100 \cite{krizhevsky2009learning}, Tiny ImageNet \cite{chrabaszcz2017downsampled} and ImageNet \cite{deng2009imagenet} datasets. For each ShuffleNet v1 and ShuffleNet v2 model, we select several group numbers and network sizes. Furthermore, we show the trade-off of our FLOP reduction and model accuracy on ResNet models. 

\subsection{Datasets and Experimental Settings}

\noindent \textbf{Dataset details.} 
The CIFAR10 and CIFAR100 datasets are small-scale datasets containing 50k training images and 10k validation images, with the size of $32\times32\times3$. CIFAR-10 contains ten categories, while CIFAR-100 contains 100 classes in total. Tiny ImageNet contains 200 classes, each of those classes have 500 training images, 50 validation images, and 50 testing images, with the size of $64\times64\times3$. ImageNet contains about 1.3M training images and 50K validation images, covering common 1K classes.

\noindent \textbf{Experiment details.} On CIFAR datasets, we set weight-decay to 5e-4, momentum to 0.9, gradient clipping at global-norm of 1, and batch size to 128 for all experiments. We train the network for 200 epochs on CIFAR-10 and 300 epochs on CIFAR-100. Following ShuffleNet \cite{zhang2018shufflenet}, we adopt the linear-decay learning rate policy for ShuffleNet and its variants. We adopt step learning rate for ResNet to keep consistent with ordinary ResNet setting. Specifically, the initial learning rates of ShuffleNet v1 models are set to 0.1, and those of ShuffleNet v2 models are set to 0.05. We set the trade-off parameter $\lambda$=0.5 for ShuffleNet v1 and $\lambda$=0.1 for ShuffleNet v2, with a warm up process in the first five epochs for both ShuffleNet v1 and v2. All experiments on CIFAR datasets are run for five times to obtain the average accuracy and their standard deviations. 

For Tiny ImageNet, we resize the $64\times64$ original images to $256\times256$, and then crop the $224\times224$ center. We use the exact hyper-parameters for training network on the CIFAR-100 dataset, except that the initial learning rate, $\lambda$, and global-norm of gradient clipping are set to  0.01, 100, and 2, respectively. 

For ImageNet, we set weight-decay to 4e-5, momentum to 0.9, and batch size to 1024 for all experiments. Following ShuffleNet \cite{zhang2018shufflenet}, we adopt the linear-decay learning rate policy and train the networks for 300000 iterations for ShuffleNet and its variants. The initial learning rates of ShuffleNet v1 models are set to 0.5. 
We set $\lambda$=0.2 and use the same hyper-parameters and training settings used in ShuffleNet \cite{zhang2018shufflenet}, with a $\lambda$ and learning rate warm up process in the first five epochs. Compared to networks on the CIFAR datasets, we set our dynamic shuffle auxiliary networks on the ImageNet dataset four times wider to obtain a stronger representation ability for large-scale datasets.



\begin{table}[t!]
	\caption{ Classification Accuracy (\%) on CIFAR-100 
	}
	\centering
    \small
	\begin{tabular}{c|c|c}
	    \hline
		Network&Manual (\%)&Dynamic (\%)\\
		 \hline
		ShuffleNet v1 ($1\times$, g=3)&70.53\textpm 0.20&\textbf{73.32}\textpm\textbf{0.22}\\
		ShuffleNet v1 ($1\times$, g=8)&70.09\textpm 0.29&\textbf{73.10}\textpm\textbf{0.28}\\
		ShuffleNet v2 (1$\times$)&71.32\textpm 0.56&\textbf{72.70}\textpm\textbf{0.25}\\
		ShuffleNet v2 (1.5$\times$)&72.74\textpm 0.26&\textbf{74.36}\textpm\textbf{0.26}\\
		\hline
	\end{tabular}
\label{table_CIFAR100}
\end{table}

\subsection{Performance Comparison Experiments} 
\noindent \textbf{Comparison with baselines.} We compare our model with ShuffleNet v1 and v2 by replacing their shuffle modules with the proposed dynamic shuffle module. 
The results are listed in Tables \ref{table_CIFAR100} to \ref{table_tiny_imagenet}, where $g$ denotes the group number of ShuffleNet. ``Manual" refers to manually setting the shuffle matrix in advance, i.e., the original ShuffleNet. It could tell from the tables that dynamic shuffle consistently improves the accuracy of ShuffleNet v1 and v2 on different datasets with different network settings, which shows that dynamic shuffle matrices work better than manually defined shuffle matrices.\\

\begin{table}[t!]
    \caption{ Classification Accuracy (\%) on CIFAR-10
	}
	\centering
    \small
	\begin{tabular}{c|c|c}
	    \hline
		Network&Manual (\%)&Dynamic (\%)\\ 
		 \hline  
		ShuffleNet v1 ($1\times$, g=3)&91.57\textpm 0.33&\textbf{93.11}\textpm\textbf{0.16}\\
		ShuffleNet v1 ($1\times$, g=8)&91.47\textpm 0.28&\textbf{92.99}\textpm\textbf{0.10}\\
		ShuffleNet v2 (1$\times$)&92.56\textpm 0.25&\textbf{93.11}\textpm\textbf{0.12}\\
		ShuffleNet v2 (1.5$\times$)&93.24\textpm 0.11&\textbf{93.52}\textpm\textbf{0.09}\\
		\hline
	\end{tabular}

\label{table_CIFAR10}
\end{table}

%
%
%
%

\begin{table}[t!]
	\centering
	\small
	\caption{{Classification Accuracy (\%) on Tiny ImageNet and Imagenet}}
	\begin{tabular}{c|c|c}
		\hline
		Network&Top 1 Acc. (\%)&Top 5 Acc. (\%)\\ 
		\hline
		\multicolumn{3}{c}{Tiny ImageNet}\\
		\hline
		ShuffleNet v1 ($1\times$, g=3)&55.27&79.76\\
		Dynamic Shuffle&\textbf{56.78}&\textbf{80.51}\\
		\hline
		\multicolumn{3}{c}{ImageNet}\\
		\hline
		ShuffleNet v1 ($1\times$, g=3)&66.34&86.74\\
		Dynamic Shuffle&\textbf{66.44}&\textbf{86.97}\\
		\hline
	\end{tabular}
	\label{table_tiny_imagenet}
\end{table}


\noindent \textbf{Comparison with other methods.} To further tell the superiority of dynamic shuffle compared with static learnable shuffle, we compare our method with AutoShuffleNet \cite{lyu2020autoshufflenet}, which  learns adaptive static shuffle strategy. The results are shown in Table \ref{table_auto}. 
The experimental setting in AutoShuffleNet is different from the original ShuffleNet as well as our method, we therefore compare the accuracy increment from the original ShuffleNet models for fair comparison. The results demonstrate that dynamic shuffle significantly outperforms AutoShuffleNet, which verifies our conjecture that the advantage of our method not only comes from learning ability but also come from dynamicity.  

We then compared the performance of our dynamic shuffle module with more other methods. Since our proposed dynamic shuffle is a channel mixture method, we adopt other methods on ShuffleNet v1 ($1\times$, g=3) by substituting the original network parts with the proposed modules of other works. The experiments are conducted on the CIFAR-100 dataset, following the exact experiment settings as our ShuffleNet baseline. We list the details of the comparison experiments with other methods below:

\begin{table}[tbp]
	\centering
    \small
	\caption{ Accuracy increment (\%) comparison with AutoShuffleNet\cite{lyu2020autoshufflenet} on CIFAR-100}
	\begin{tabular}{c|c|c}
        \hline
		Network&AutoShuffleNet (\%)&Dynamic Shuffle (\%)\\ 
         \hline
		ShuffleNet v1($1\times$, g=3)&1.73&\textbf{2.79}\\
		ShuffleNet v1($1\times$, g=8)&1.24&\textbf{3.01}\\
		ShuffleNet v2 1$\times$&0.65&\textbf{1.38}\\
		ShuffleNet v2 1.5$\times$&0.66&\textbf{1.62}\\
		\hline

	\end{tabular}
\label{table_auto}
\end{table}

\begin{table}[t!] 
	\centering
	\small
	\caption{{Comparisons with other dynamic or channel mixture methods on ShuffleNet v1 ($1\times$, g=3) for CIFAR-100}}
	\begin{tabular}{c|c|c|c}
		\hline
		Module &Acc. (\%) &Params& Type \\ 
		\hline
		Manual shuffle (baseline)&70.53\textpm 0.20&1.01M&channel-mix\\
		SENet \cite{hu2018squeeze}&70.88\textpm 0.12&1.05M&auxiliary\\
		MicroNet \cite{li2021micronet}&72.19\textpm 0.33&1.48M&channel-mix\\
		FLGC \cite{wang2019fully}&70.61\textpm 0.42&1.89M&conv-layer\\
		DGC \cite{su2020dynamic}&73.12\textpm 0.47&3.72M&conv-layer\\
		DyReLU \cite{chen2020dynamic}&70.49\textpm 1.12&2.56M&auxiliary\\
		ACON \cite{ma2021activate}&63.17\textpm 0.73&1.04M&auxiliary\\
		WeightNet \cite{ma2020weightnet}&70.53\textpm 0.30&1.08M&conv-layer\\
		CondConv \cite{yang2019condconv}&70.66\textpm 0.58&2.83M&conv-layer\\
		DGConv \cite{zhang2019differentiable}&69.86\textpm 0.62&1.87M&conv-layer\\
		Dynamic shuffle (ours)&\textbf{73.32\textpm 0.29}&1.09M&channel-mix\\
		\hline
	\end{tabular}
	\label{table_CIFAR100_dynamic}
\end{table}

\begin{table}[t!]
	\centering
	\small
	\caption{{Comparison of Different ResNet Variants on CIFAR-100}} 
	\begin{tabular}{c|c|c|c}
		\hline
		Network&Acc. (\%)&Params&GFLOPs \\ 
		\hline
		ResNet-50 &76.83\textpm 0.60&23.7M&1.305\\
		ResNet-50-duplicate&75.57\textpm 0.22&18.7M&1.032\\
		ResNet-50-static&76.78\textpm 0.16&18.7M&1.032\\
		ResNet-50-dynamic&76.67\textpm 0.46&19.4M&1.033\\
		ResNet-50-static-dynamic&\textbf{77.68\textpm 0.25}&19.5M&1.033\\
		\hline
	\end{tabular}
	\label{table_resnet}
\end{table}

\begin{itemize}
\item{SE-Net\cite{hu2018squeeze}: we apply the SE module directly on the original shuffled features of ShuffleNet.}
\item{Micro-Net\cite{li2021micronet}: we use the Dynamic Shift-Max module of Micro-Net to substitute the shuffle operation, which is proposed to dynamically enhance group-wise channel mixture, similar to our dynamic shuffle.}
\item{FLGC\cite{wang2019fully}/Dynamic GC\cite{su2020dynamic}: we implement the FLGC (Fully Learnable Group Convolution)/Dynamic GC (Dynamic Group Convolution) module to substitute both the group convolution layer and shuffle layer, since FLGC/Dynamic GC modules serve as learned static/dynamic group convolution strategy respectively.}
\item{Dynamic ReLU\cite{chen2020dynamic}/ACON\cite{ma2021activate}: we change all the ReLU functions into dynamic ReLU/ACON. Since its dynamic nature, the ACON training process is unstable, and the accuracy has dropped. We will do further finetune upon its default settings to alleviate this.}
\item{WeightNet\cite{ma2020weightnet}: We change the normal and depth-wise conv layers into WeightNet conv layers.}
\item{CondConv\cite{yang2019condconv}: We change the point-wise conv layers into CondConv2D layers.}
\end{itemize}
All other settings of comparison experiments are the same with those of ShuffleNet v1 (g=3) experiments on CIFAR-100. Results are shown in Table \ref{table_CIFAR100_dynamic}. Here ``Type" indicates the methods' motivations, where ``channel-mix" denotes small modules applicable on all channel mixture scenes, ``conv-layer" denotes designing whole layers as new network architectures, and ``auxiliary" denotes enhancing network performance with plug-in modules. We can see that, dynamic shuffle significantly outperforms all the other methods. Compare to manual method (baseline) or learnable  pre-define methods (FLGC),  we show that shuffle according to the input is more powerful.
{Compare to the other dynamic networks, our method still has advantageous representation ability.}


\subsection{Computation Cost Reduction Experiments on ResNet} 
We show the static-dynamic-mix shuffle's effect on reducing computation cost by conducting experiments on CIFAR-100 with ResNet-50, where the $1\times1$ convolution expansion layer is substituted by different shuffle matrices. 
For ResNet-50-duplicate, the $1\times1$ convolution layer is substituted with stacked identity matrices, i.e., duplicating the feature maps for expansion. For ResNet-50-static, only learnable static shuffle matrices are used, and for ResNet-50-dynamic, only dynamic matrices are applied. ResNet-50-static-dynamic  refers to using the static-dynamic-mix shuffle matrices. 

The results are shown in Table \ref{table_resnet}. Although replacing the $1\times1$ convolution expansion layer with simple duplication reduces the computation significantly, the accuracy also drops significantly because of the lack of feature map diversity. Shuffling the feature maps with learnable static or dynamic permutation matrix for expansion increases the representation ability and restores the accuracy to a little lower than original ResNet with almost the same FLOPs as duplication. Benefiting from the representation ability of dynamic shuffle and the training stability of static shuffle, the proposed static-dynamic-mix strategy achieve a surge in the final accuracy.





\begin{table*}[t!]
	\centering
    \small
	\caption{Ablation experiment results on CIFAR-100. ``Binarization" indicates the binarizing process of the shuffle matrix. ``Orth. regularization" is the constraint to make the shuffle matrix orthogonal.}
    \setlength{\tabcolsep}{1mm}{
	\begin{tabular}{c|c|c|c|c}
	   \hline
        Network&Binarization&Orth. regularization&Dynamic matrix&Acc. (\%)\\ 
         \hline
		dynamic shuffle &\Checkmark&\Checkmark&\Checkmark&\textbf{73.32}\\
		w/o binarization&\XSolidBrush&\Checkmark&\Checkmark&69.90\\
		w/o orth. regularization&\Checkmark&\XSolidBrush&\Checkmark&55.95\\
		w/o dynamic input&\Checkmark&\Checkmark&\XSolidBrush&71.26\\
        \hline
	\end{tabular}}
\label{table_ablation}
\end{table*}

\begin{table*}[t!]
	\centering
	\small
	\caption{The comparison between two permutation matrices generation method.``Full Channel " indicates only use two bigger matrices to generation permutation matrices. ``Sharing" is our permutation matrices sharing method. ``Baseline Params" is the  parameter numbers of baseline networks. ``Params.  $\uparrow$ (\%)" is the additional parameter number of the auxiliary network and its corresponding percentage.}
	\begin{tabular}{c|c|c|c|c|c}
		\hline
		\multirow{2}{*}{Network}&\multirow{2}{*}{Baseline Params.}&\multicolumn{2}{c|}{Full Channel}&\multicolumn{2}{c}{Sharing}\\
		\cline{3-6}
		&&Params. $\uparrow$ (\%)&Acc.&Params. $\uparrow$ (\%)&Acc. \\
		\hline
		\multicolumn{6}{c}{CIFAR-10}\\
		\hline
		ShuffleNet v1 (1$\times$, g=3) &0.93M&0.14M (15.1)&93.02&0.08M / 8.4\%&\textbf{93.11}\\
		ShuffleNet v1 (1$\times$, g=8) & 0.93M&0.35M (37.8)&92.65&0.07M / 7.0\%&\textbf{92.99}\\
		ShuffleNet v2 (1$\times$) & 1.27M&0.51M (40.1)&93.00&0.13M / 10.5\%&\textbf{93.11}\\
		ShuffleNet v2 (1.5$\times$) & 2.49M&1.15M (46.7)&\textbf{93.58}&0.30M / 12.0\%&93.52\\
		\hline
		\multicolumn{6}{c}{CIFAR-100}\\
		\hline
		ShuffleNet v1 (1$\times$, g=3) & 1.01M&0.14M (13.9)&\textbf{73.44}&0.08M (7.7)&73.32\\
		ShuffleNet v1 (1$\times$, g=8) & 1.07M&0.35M (32.9)&72.66&0.07M (6.1)&\textbf{73.10}\\
		ShuffleNet v2 (1$\times$) & 1.36M&0.51M (37.7)&72.59&0.13M (9.8)&\textbf{72.70}\\
		ShuffleNet v2 (1.5$\times$) & 2.58M&1.15M (44.5)&\textbf{74.99}&0.30M (11.6)&74.36\\
		\hline
	\end{tabular}
	\label{table_param}
\end{table*}

\subsection{Ablation Studies}\label{ablation}
For a better understanding of the proposed method, we investigate the effectiveness of each component by  conducting ablation experiments on CIFAR-10 and CIFAR-100 with ShuffleNet v1 and ShuffleNet v2.




\textbf{The effectiveness of permutation matrices sharing:}  The experiment settings and results are summarized in Table \ref{table_param}. ``Sharing" denotes our permutation matrices sharing strategy. ``Full Channel" denotes where the auxiliary network straightly uses the Kronecker product result of two bigger matrices as the permutation matrices. Although the ``Full Channel" strategy uses 2 to 4 fold additional parameters than our permutation matrices sharing method, the sharing method has merely shown slight decrease, even increase, in many experiments. The result shows that the combination of matrix sharing and the manual shuffle  is a good enough constrained paradigm for generating dynamic permutation matrices.

\textbf{The effectiveness of binarization, orthogonality regularization, and dynamic input:} We conduct ablation experiments on CIFAR-100 with ShuffleNet v1, by removing binarization, orthogonality regularization, and dynamic input, respectively. The experiment settings and results are summarized in Table \ref{table_ablation}. When removing the binarization operation, the auxiliary network actually plays a role to produce the weights of a learnable  $1\times1$ convolution. Although it carries more information than a binarized one, dynamic shuffle surprisingly outperforms it. 
This is because with orthogonality regularization and softmax, the final desired matrix is binarized and adding binarization makes the optimization of the desired matrix easier. Binarization can also be considered as adding more constraints to reduce the optimization space, and thus avoid overfitting on small datasets. 
We have to emphasize that, without binarization, matrix multiplication would be necessary to be implemented on networks for inference, but our method with binarization can be performed by memory shifting. We also conduct experiments without training binarization but with testing binarization, during which the network only produces random classification results. This shows that the binarization operation is necessary to obtain useful shuffle matrices.

\begin{figure*}
	\centering
	\small
	\subfigure[ShuffleNet v1 ($1\times$, g=3)]{
		\begin{minipage}[t]{0.23\textwidth}
			\centering
			\includegraphics[width=0.95\textwidth]{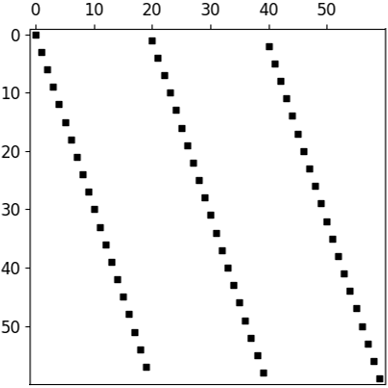}
		\end{minipage}
	}
	\subfigure[Dynamic Shuffle]{
		\begin{minipage}[t]{0.23\textwidth}
			\centering
			\includegraphics[width=0.95\textwidth]{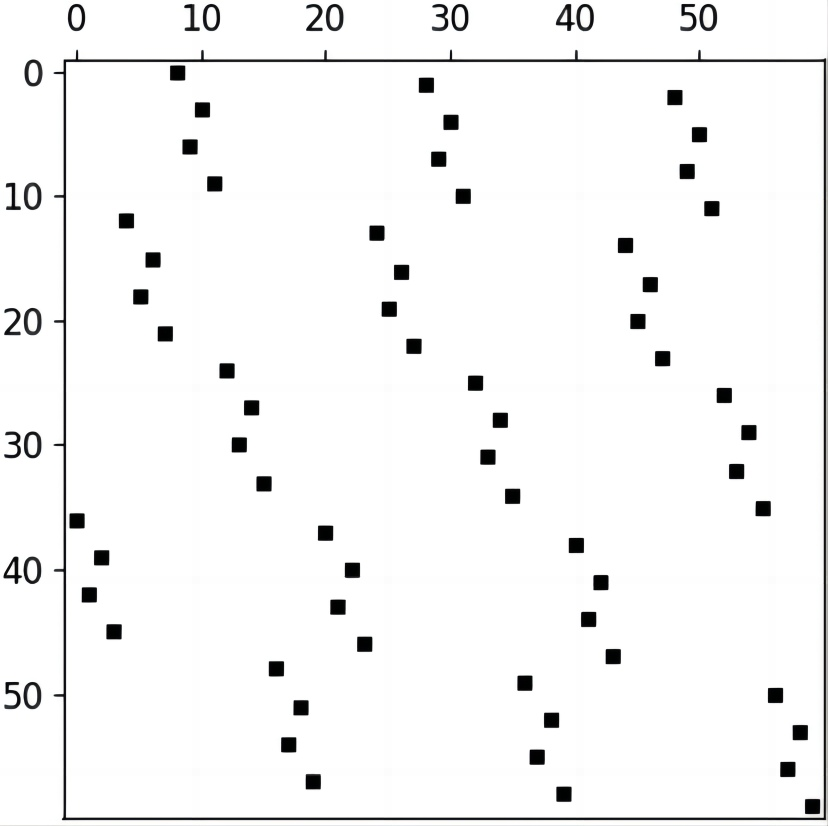}
		\end{minipage}
	}
	\subfigure[AutoShuffleNet v1 ($1\times$, g=3)]{
		\begin{minipage}[t]{0.23\textwidth}
			\centering
			\includegraphics[width=0.95\textwidth]{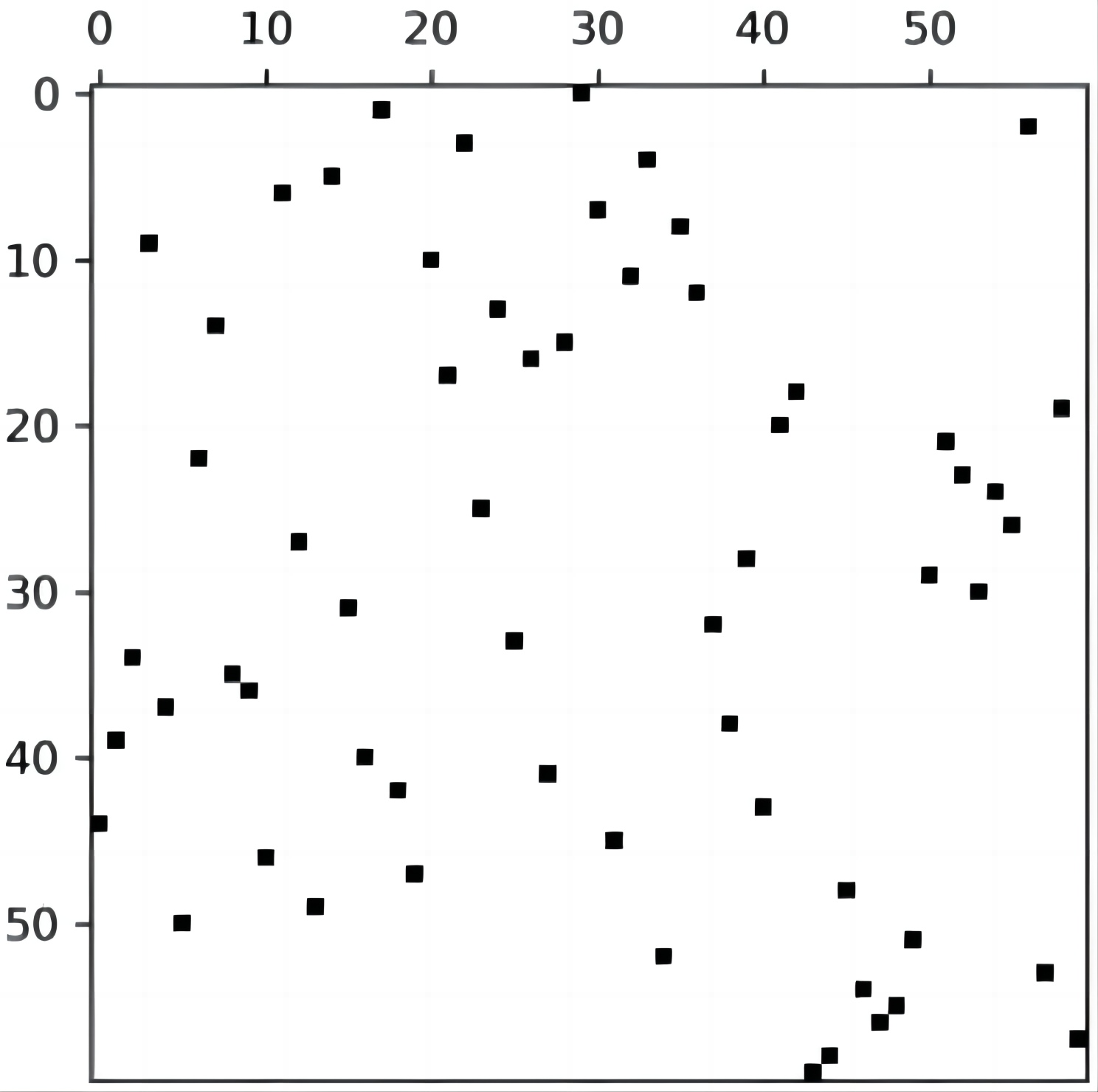}
		\end{minipage}
	}  
	\subfigure[Ablation ``w/o orth loss"]{
		\begin{minipage}[t]{0.23\textwidth}
			\centering
			\includegraphics[width=0.95\textwidth]{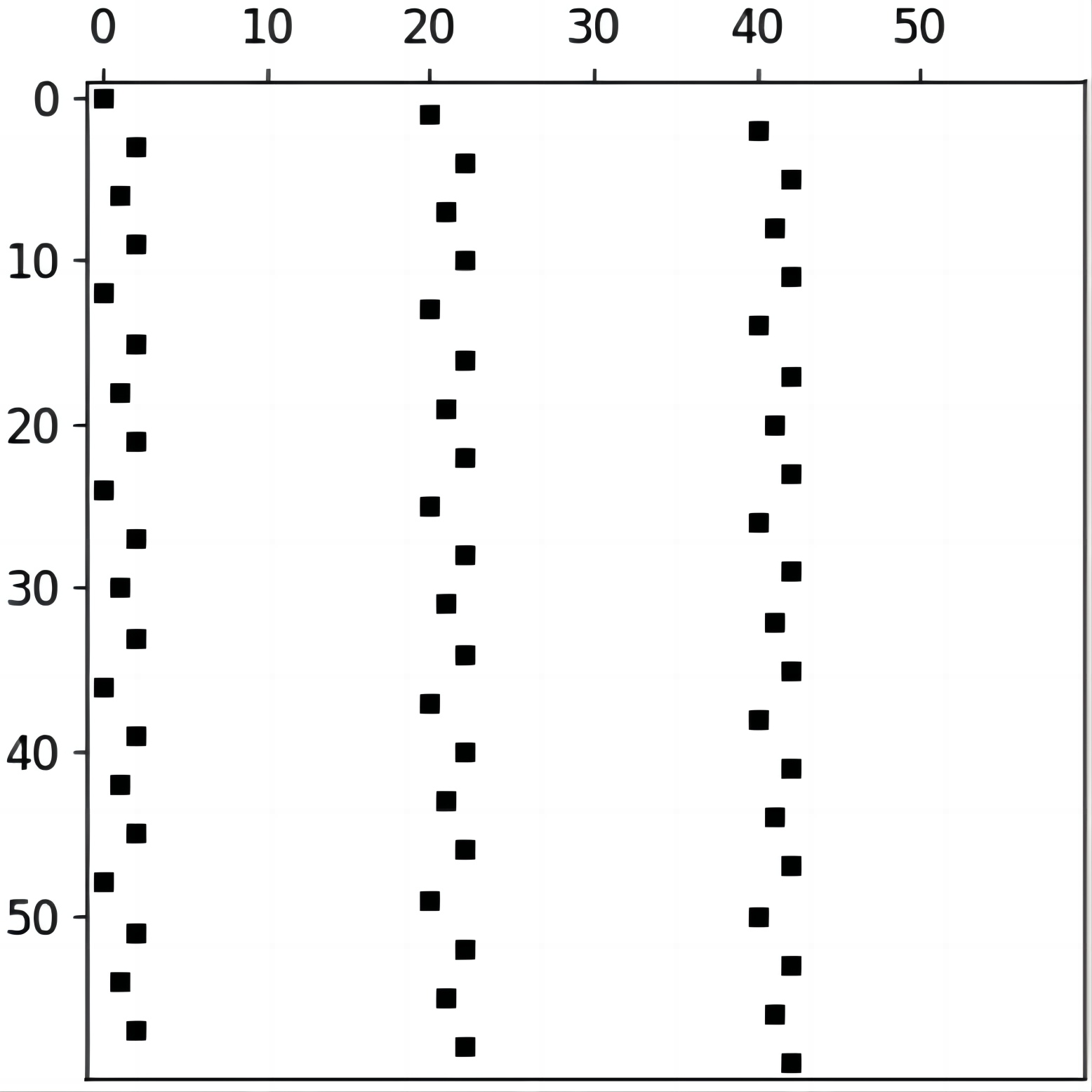}
		\end{minipage}
	}  
	\caption{The sampled binary shuffle matrices of the first shuffle unit from different networks. Each graph represents a $60\times60$ matrix, with the black dots indicating the entries of 1's}
	\label{visfig}
\end{figure*}

Without orthogonality regularization, the desired vector generated by softmax is not one-hot, and directly binarizing it will lose too much useful information. In this case, some channels are repeated, and some are dropped after shuffling, so the result is poor. The visualization in Figure \ref{visfig} (d) also shows that the learned matrix only selects part of the channels.

Removing dynamic input means the two small matrices are learned directly without input. With the orthogonal constraint, the shuffle matrix is a permutation matrix, and there is no channel information loss after shuffling. Thus, the result is significantly better than that of removing the orthogonality regularization. However, we can see that the result is poorer than the proposed dynamic shuffle, which is because decomposing the shuffle matrix into Kronecker product of small matrices significantly reduces the optimization space, and it is hard to obtain a good result in the reduced space. Nevertheless, when introducing dynamic input, the increase of representation ability compensates for this drawback.

\begin{figure*}[h!]
	\centering
	\small
	\subfigure[Total loss and accuracy]{
		\begin{minipage}[t]{0.44\textwidth}
			\centering
			\includegraphics[width=\textwidth]{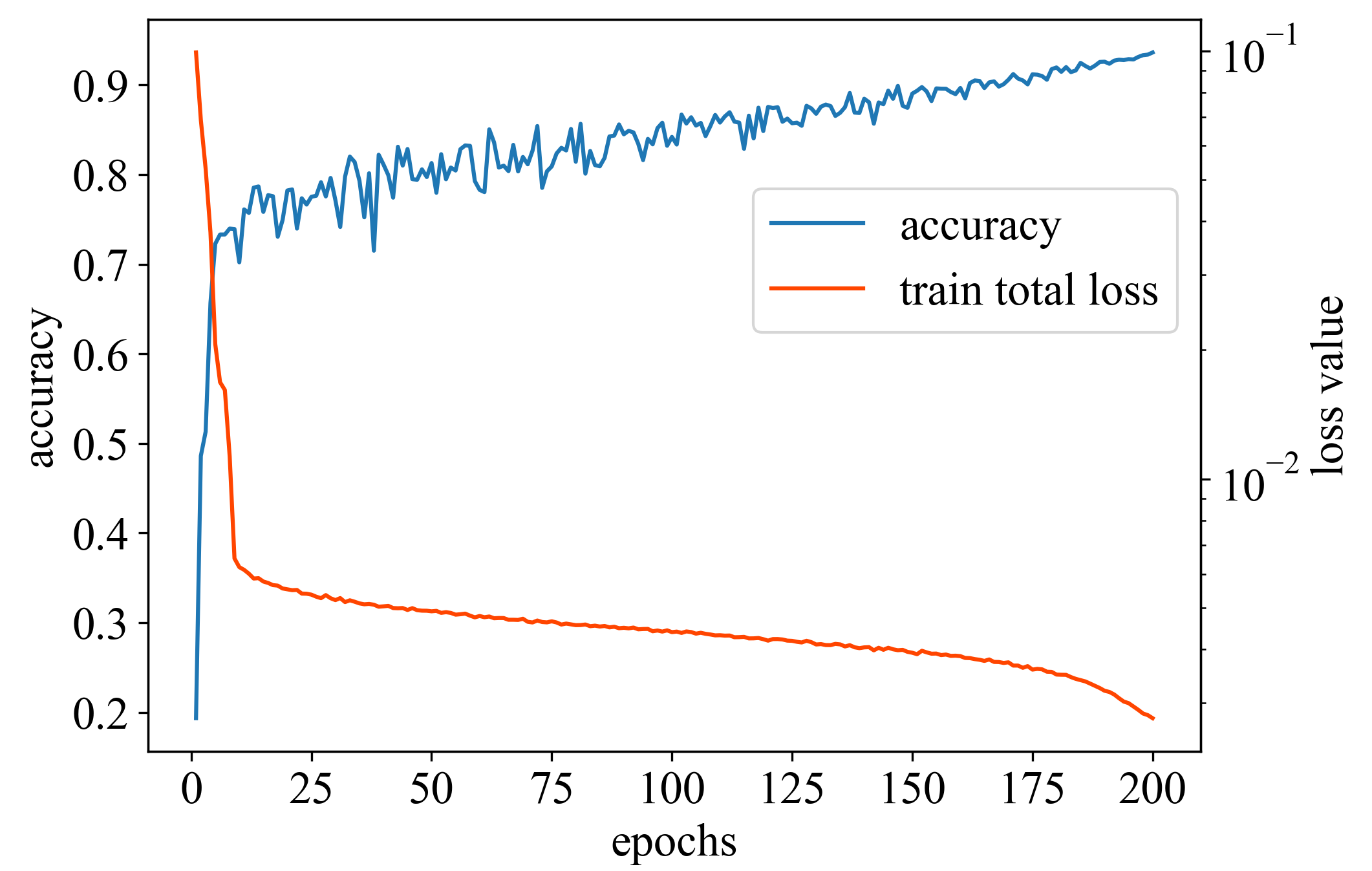}
		\end{minipage}
	}
	\subfigure[CE loss comparison]{
		\begin{minipage}[t]{0.44\textwidth}
			\centering
			\includegraphics[width=\textwidth]{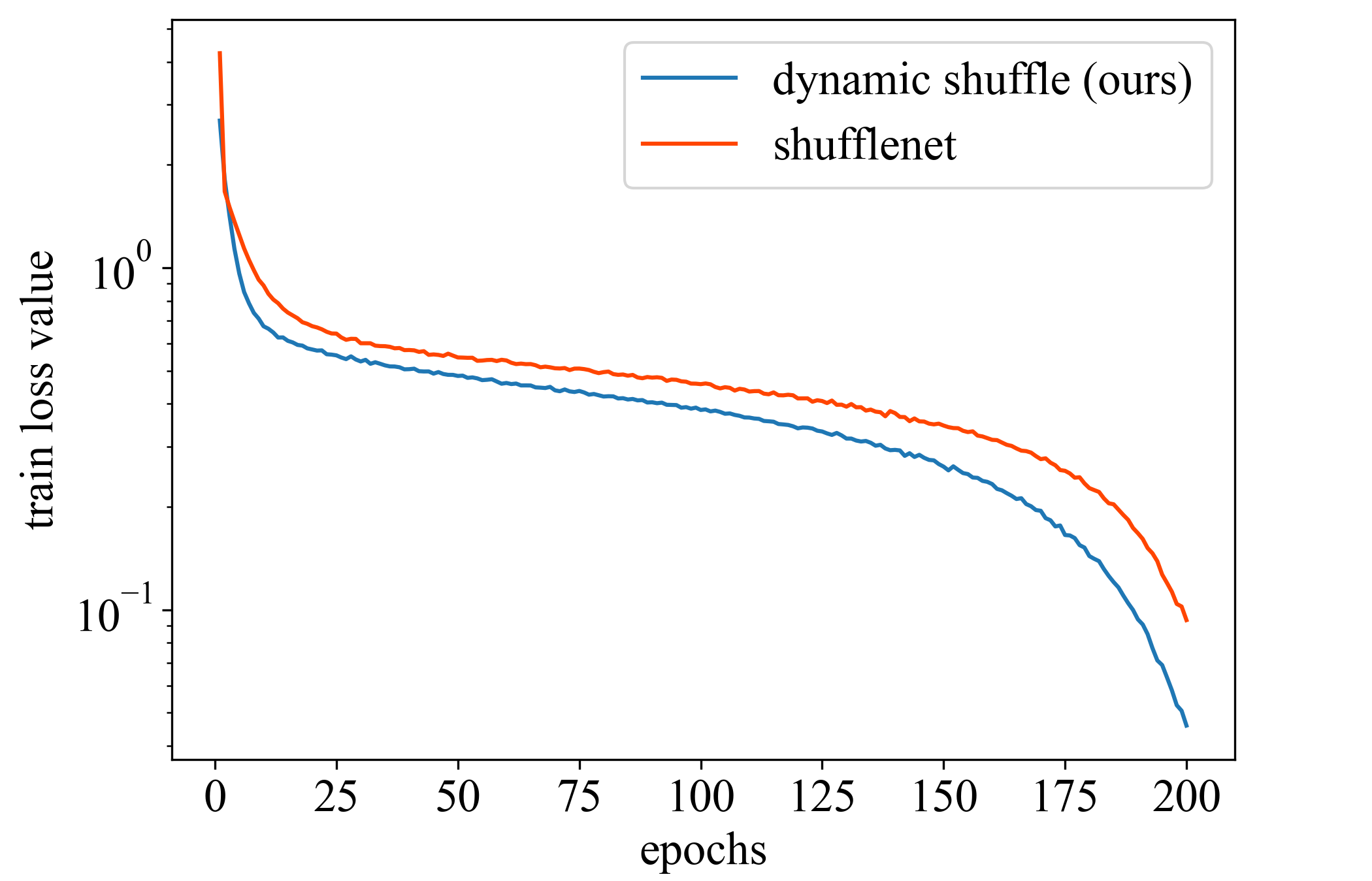}
		\end{minipage}
	}
	\caption{ Convergence curves on CIFAR-10. }
	\label{cvrfig}
\end{figure*}

\subsection{Shuffle Matrix Visualization}
To provide a more intuitive comparison of the shuffling operation among different methods, we visualize the shuffle matrices of manual shuffle, dynamic shuffle, automatic shuffle, and dynamic shuffle without orthogonal loss regularization. We show the first shuffle unit in ShuffleNet v1 ($1\times$, g=3) of these methods in Figure \ref{visfig}.
Since our shuffle matrices are dynamically generated, we select the first generated matrix during the test stage of the final training epoch.

The shuffle matrix shown in Figure \ref{visfig} (a) of ShuffleNet has shown the predefined formality, which is not the optimal shuffle strategy. Figure \ref{visfig} (b) shows a shuffle matrix of our dynamic shuffle model. The reoccurring patterns in the graph are due to the characteristics of the Kronecker product, that our shuffle matrix is the product of four small matrices, determining the global, local, sharing, and manual shuffle strategies respectively, as indicated in Eq. (\ref{four_mat}). This characteristic reduces the optimization space and helps the convergence of the dynamic network training. Figure \ref{visfig} (c) shows a learned static shuffle matrix from AutoShuffleNet, which applies for all input features to the corresponding layer, mainly focusing on optimizing the inherent structure of the network.  Figure \ref{visfig} (d) shows the sampled shuffle matrix in the ablation experiment removing orthogonality regularization, in which the dynamic auxiliary network could not converge and only select part of the channels.



\subsection{Convergence Analysis}
The dynamic network actually makes the network a high-order function to improve the network representation ability. Since the relationship between input and output is more complex, the training process may become unstable. Here we study the converge of training dynamic shuffle. We illustrate the convergence curves of dynamic shuffle and ShuffleNet v1 ($1\times$, g=3) on CIFAR-10 in Figure \ref{cvrfig}. It can be seen that, dynamic shuffle converges well. Specifically, it converges a bit faster than ShuffleNet from the beginning of the experiment, and achieves lower loss value in the final epochs. 


\section{Conclusion}
In this paper, we extend ShuffleNet to dynamic shuffle, which brings a considerable accuracy increase with nearly no extra computation. We also show that the dynamic shuffle module can be used to directly substitute the $1\times1$ convolution layer in bottleneck modules, sparing huge memory and computation cost with even an accuracy increase. We expect our dynamic shuffle module used in future models as a channel mixture alternative. We will consider dynamic shuffle in conjunction with other network compression methods like pruning and quantization in the future.


\bibliographystyle{elsarticle-harv}
\bibliography{Dynamic_Shuffle_A_Channel_Mixture_Method}



\end{document}